%% file: Active_suspension.tex
\documentclass{article}
\usepackage[left=2cm, right=2cm, top=2cm]{geometry}

\usepackage[utf8]{inputenc}
\usepackage{hyperref}
\hypersetup{
    colorlinks=true,
    linkcolor=blue,
    filecolor=magenta,      
    urlcolor=blue,
    pdftitle={Overleaf Example},
    pdfpagemode=FullScreen,
    }

\usepackage{times}
\usepackage{latexsym}
\usepackage{subfig}
\usepackage[T1]{fontenc}
\usepackage{authblk}
\usepackage{microtype}
\usepackage{algorithm}
\usepackage{algorithmicx}  
\usepackage{algpseudocode}
\usepackage{amsmath,amssymb,amsfonts}
\usepackage{textcomp}
\usepackage{gensymb}
\usepackage{textgreek}
\usepackage{xcolor}
\usepackage{graphicx}
\usepackage{multirow}
\usepackage{siunitx}
\usepackage{tikz} 
\usepackage{pgfplots}
\pgfplotsset{compat=newest} 
\usepgfplotslibrary{units}
\graphicspath{{Images/}}

\include{defs}

\title{\textbf{Autonomous Control of a Novel Closed Chain Five Bar Active Suspension via Deep Reinforcement Learning}}
\author[1]{Nishesh Singh}
\author[2]{Sidharth Ramesh}
\author[1]{Jyotishka Duttagupta}
\author[1]{Abhishek Sankar}
\author[3]{Leander Stephen D'Souza}
\author[4]{Sanjay Singh}
\affil[1]{Department of Mechanical Engineering, Manipal Institute of Technology}
\affil[2]{Department of Computer Science Engineering, Manipal Institute of Technology}
\affil[3]{Department of Mechatronics, Manipal Institute of Technology}
\affil[4]{Department of Information and Communication Technology, Manipal Institute of Technology}
\date{June 2021}
\begin{document}
 \maketitle
\section{Abstract}
 Planetary exploration requires traversal in environments with rugged terrains. In addition, Mars rovers and other planetary exploration robots often carry sensitive scientific experiments and components onboard, which must be protected from mechanical harm. This paper deals with an active suspension system focused on chassis stabilisation and an efficient traversal method while encountering unavoidable obstacles. Soft Actor-Critic (SAC) was applied along with Proportional Integral Derivative (PID) control to stabilise the chassis and traverse large obstacles at low speeds. The model uses the rover's distance from surrounding obstacles, the height of the obstacle, and the chassis' orientation to actuate the control links of the suspension accurately. Simulations carried out in the Gazebo environment are used to validate the proposed active system.
 \newline
 \newline
 \textbf{Keywords}: Deep Reinforcement Learning, Soft Actor-Critic, Active suspension, Mars Exploration Rover (MER), Five bar mechanism, Space robotics, OpenAI Gym, \texttt{gym-gazebo} 
 
\section{Introduction}
Locomotion over rough terrain remains a formidable challenge for mobile robots. Never much more so than in the sphere of planetary exploration. In just the past few years, mobile robots are being employed at an ever-increasing rate to explore the surfaces of numerous extraterrestrial bodies \cite{enwiki:1028258768}\cite{enwiki:1028824573}. The robot provides mobility to onboard scientific instruments, which can be carried to specific objective sites, enabling scientists to gather valuable information and conduct research. Hence, increased traversal capabilities are essential for the robot to navigate the environment and accomplish these tasks. \par

NASA's rocker-bogie system is the most prominent of the several mobile systems that have been proposed \cite{harrington2004challenges}. It was first used on the Mars rover Sojourner and over time became the suspension of choice for Mars exploration vehicles, given its superior stability and obstacle climbing ability. The rocker-bogie suspension, which features a lateral differential mechanism, enables a six-wheeled rover to passively keep all six of its wheels in contact with the ground while traversing across severely uneven and rugged terrain \cite{1571187}. This feature has two crucial advantages. First, it equilibrates the pressure of the wheels on the ground, and second, it helps to nominally keep the wheels in contact with the ground while climbing over uneven obstacles \cite{harrington2004challenges}.

\begin{figure}[htbp]
\centering
\includegraphics[width=10cm]{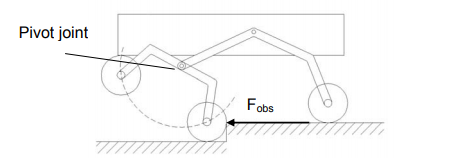}
\caption{The bogie overturn problem \cite{barlas2004design} describes a greater than 90 \degree revolution of the bogie about the bogie-rocker revolute joint, which leads to a collapse of the mechanism and renders the rover immobile.}
\label{fig:bogie-overturn}
\end{figure} 

\begin{figure}[htbp]
\centering
\includegraphics[width=10cm]{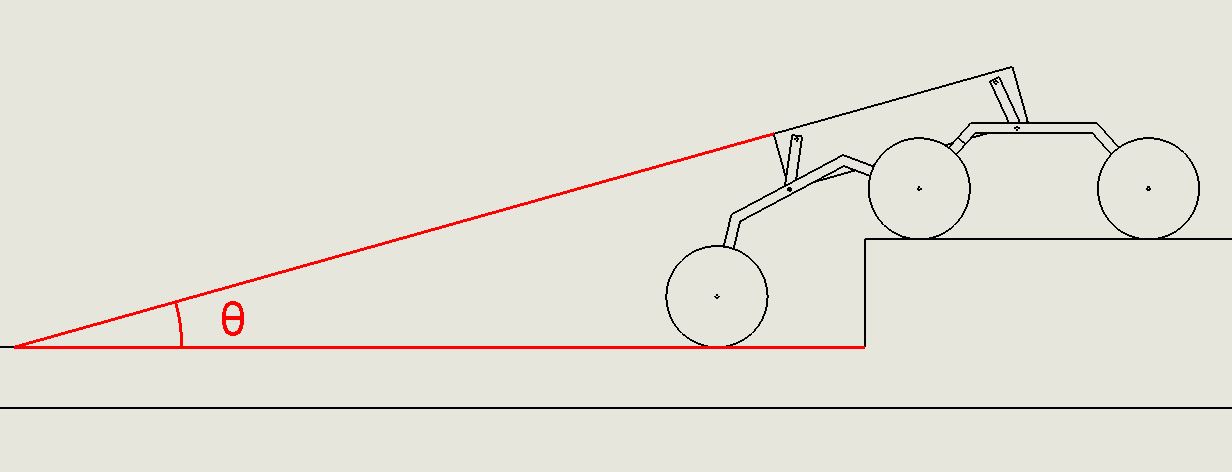}
\caption{Pictorial representation of the chassis pitch. Pitch is defined as the angle subtended between the longitudinal axis of the chassis and the ground plane.}
\label{fig: chassis pitch}
\end{figure}

The bogie link is a centrally pivoted link with a wheel mounted on its two ends. A problem commonly associated with the rocker-bogie suspension is bogie overturn \cite{barlas2004design}. It is a term used to describe a greater than 90 \degree  revolution of the bogie about the bogie-rocker revolute joint, which essentially flips the bogie link and renders the rover immobile. Such a situation cannot be corrected remotely and requires manual intervention to bring the rover out of its stranded state. The subjection of the suspension to extreme traversal situations such as descending steep drops, colliding head-on with an obstacle at high speed, and the middle wheel getting lodged on an obstacle, can result in bogie overturn. By manipulating the design of the suspension, one can avoid the occurrence of bogie overturn. Solutions include manipulating the geometry of the linkage mechanism, usage of mechanical stops, and generally operating at slower speeds. All these solutions significantly reduce the mobility of the suspension and the speed of the rover. This work attempts to develop a solution by collecting, identifying and presenting a set of desirable characteristics for a mobile wheeled rover suspension. The objective for determining these characteristics is to evaluate the performance of a system and provide reviewers with a tool to compare the performance of a new system impartially concerning the present state of the art. \newline
Characteristics for a  suspension system to be used for navigating and exploring uneven terrain, especially in operations that are beyond the scope of manual intervention, are identified as follows:

\begin{itemize}
  \item The suspension should be able to preserve the chassis's lateral and longitudinal stability at all times, even when negotiating large obstacles and varied terrain \cite{barlas2004design}. The longitudinal stability is a direct measure of the pitch (see Figure \ref{fig: chassis pitch}) of the chassis and, lateral stability is a direct measure of the roll.
  \item The suspension must possess increased traversability, which refers to a rovers ability to traverse uneven and irregular terrain while ensuring ground contact on as many wheels as possible \cite{apostolopoulos2001analytical}. This characteristic is of paramount importance in the scope of the present work. 
  \item The suspension should be capable of changing its heading on command and navigate through cluttered environments. This property is referred to as the manoeuvrability of a mobile robot \cite{apostolopoulos2001analytical}. 
  \item The suspension must have greater trafficability, which refers to the ability of a rover to generate traction and overcome resistances \cite{apostolopoulos2001analytical}. 
  \item The suspension must be able to scale obstacles several times its tire radius. Usual consumer Automobiles can only climb obstacles with height less than or equal to the radius of the tire \cite{barlas2004design}. 
  \item The suspension must absorb and offload any impact loads the rover might experience while navigating the terrain. 
  \item The suspension mechanism must not have positions in its motion that might render the rover immobile, such as mechanical singularities or cases such as bogie-overturn in the rocker-bogie mechanism.
\end{itemize}
This work outlines a proposal for a novel suspension system, considering the challenges highlighted with the current state of the art and the performance evaluation indices outlined above.

\begin{figure}[htbp]
\centering
\includegraphics[width=10cm]{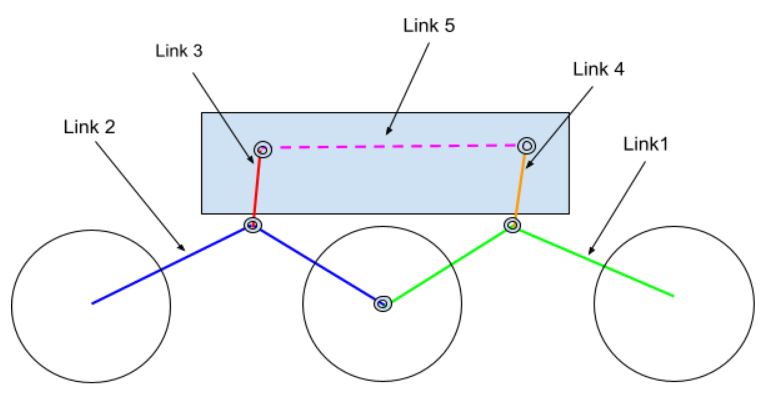}
\caption{Kinematic diagram of the proposed planar five bar mechanism}
\label{fig: 5bar-line-diagram}
\end{figure}

The new suspension system is meant to be used for a wheeled robot. It is based on a modified five-bar mechanism \cite{campos2010development} that has outer wheels mounted to extrapolated sections of links 1 and 2, and the middle wheel mounted at the revolute joint between links 1 and 2 as shown in Fig. \ref{fig: 5bar-line-diagram}. The wheel secured on the extrapolated section of link 1 is henceforth referred to as the front wheel. Each side of the suspension consists of two bogies, two vertical links 3 and 4 (henceforth referred to as control links), and three wheels. A laterally mirrored setup on the other side of the chassis completes the six-wheeled suspension setup.The suspension has all six wheels powered, making it a six-wheel drive. The control links are mounted to the chassis, which forms the fifth Link. The closed chain planar five-bar linkage mechanism forms the basis of the suspension. \newline
The proposed suspension system is not structurally stable in its present configuration as the 5 bar mechanism cannot retain its structural integrity under loading. Thus the use of a mechanism to constrain the five-bar mechanism to the chassis becomes necessary. One can achieve the same using either passive or active methods. \par
In passive systems, the mechanical configuration of the linkages itself is responsible for the interaction of the wheels with the terrain. No complex control systems dictating the position of the links or the reaction of the wheels with the ground is required. A passive suspension system typically comprises of a combination of shock absorbers and springs. It is characterised by an inability to actively alter suspension parameters (such as suspension stiffness, damping coefficient, suspension travel, geometry and more.) in response to different road/terrain conditions \cite{Sun2004activesuspension}. A practical method to constrain the five-bar mechanism to the chassis is to restrict the motion of the turning pairs between (see Fig. \ref{fig: 5bar-line-diagram}) links 3 and 5 and links 4 and 5 using torsion springs. Torsion springs' rate is defined in N-mm per deg, i.e. it provides increasing torque in response to increasing angular displacement. This setup provides a simple, cost-effective, and implementable method of constraining the 5 bar mechanism. In addition to this, it confers shock absorption properties onto the suspension.

\begin{figure*}[htbp]
\begin{tabular}{cccc}
\subfloat{\includegraphics[width = 3.2in]{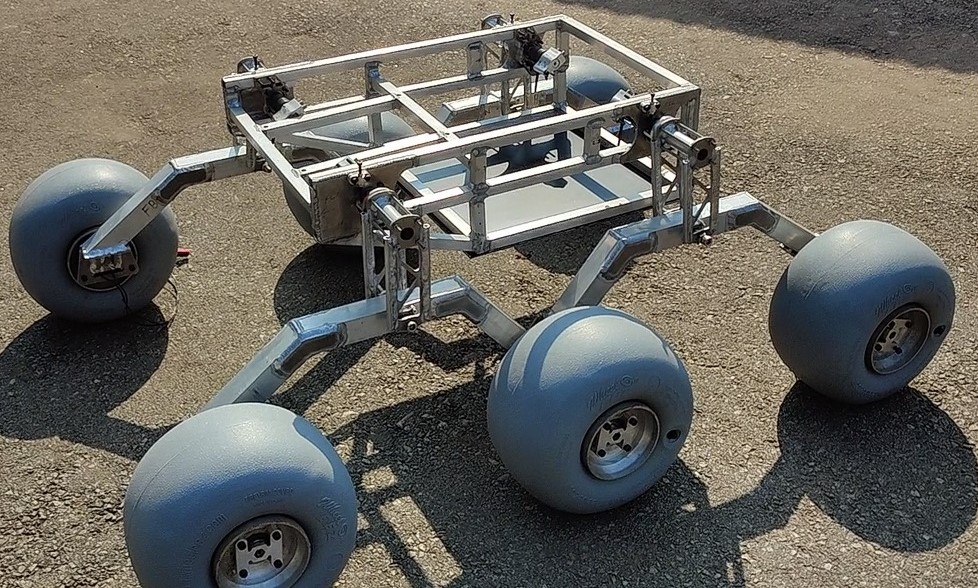}} &
\subfloat{\includegraphics[width = 3.45in]{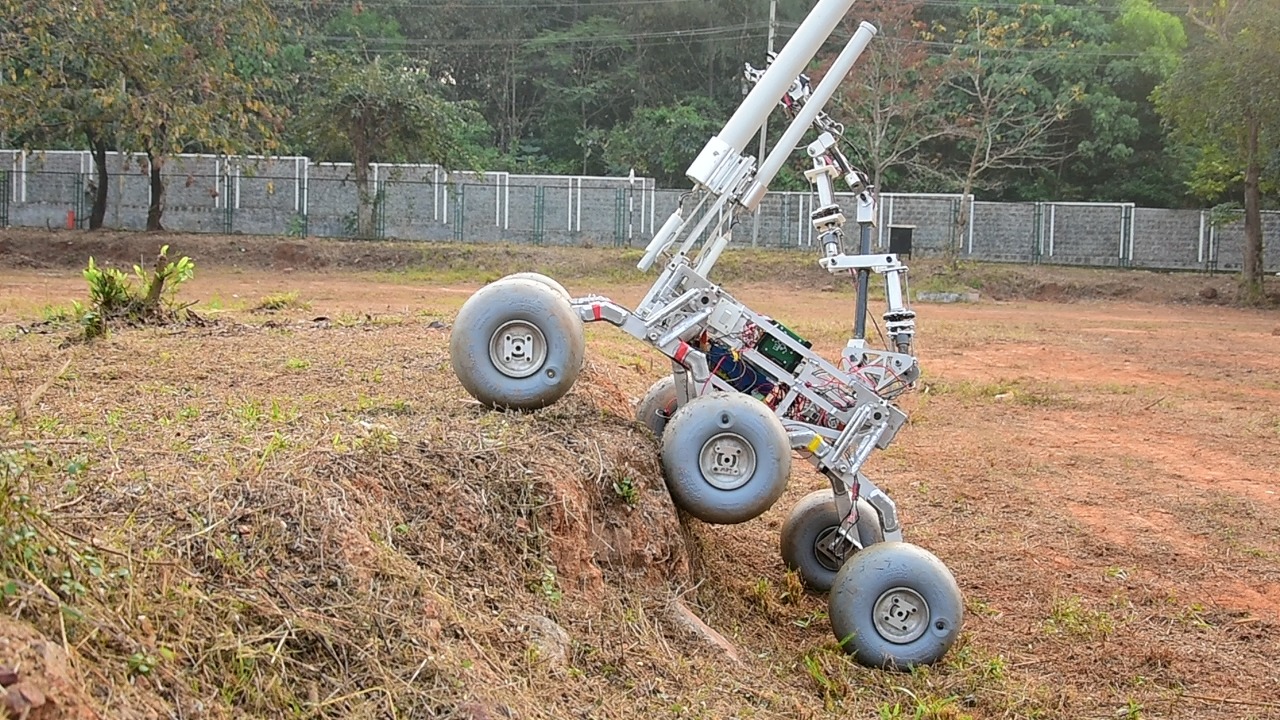}}
\end{tabular} 
\caption{Prototype rover comprising of a torsion spring constrained passive five bar suspension system.}
\label{rover-image}
\end{figure*}

Compared to passive systems, active systems present significantly higher levels of complexity concerning the adaption of the wheels with the ground. However, unlike purely passive suspension systems, an active system provides a high degree of control over the reconfiguration capabilities of the suspension geometry. Thus, it overall confers onto the rover, more remarkable traversability. Furthermore, an active system makes use of actuators to change and modify the kinematics of the suspension. Not only can it achieve wheel adaption to the ground but also actively reconfigure the geometry to aid in traversability by, for example, increasing traction on the ground . An active control solution for the same entails controlling the rotation of the turning pairs between (refer to Figure \ref{fig: 5bar-line-diagram}) links 3 \& 5, and links 4 \& 5 using rotary actuators.

Considering the different types of obstacles that appear in various unknown terrains, the control method for the suspension must be adaptable and able to make decisions without having seen its exact surroundings at any last moment. It cannot be assumed that the rover has access to all the required details of its surroundings, making it essential that the control method is proactive in deciding to traverse an obstacle rather than simply reactive to its current state. Keeping in mind the proactivity of the state, we chose deep reinforcement learning as the method of controlling the suspension. In the field of robotics, Deep Reinforcement Learning is currently seen as one of the most promising avenues of research for an end to end automation of robots. \cite{pmlr-v87-mahmood18a}. 

Reinforcement Learning is a form of machine learning different from supervised and unsupervised learning, where an intelligent agent learns to take a series of actions by maximising a cumulative reward in order to extrapolate or generalise to situations not present in the training set \cite{10.5555/3312046}. While there are many approaches to solving reinforcement learning problems, this study mainly focuses on using policy gradient methods that perform better in continuous action spaces. In particular, an off-policy algorithm Soft Actor-Critic (SAC), introduced in \cite{pmlr-v80-haarnoja18b} and improved in \cite{haarnoja2019soft} is used to predict the actions the rovers control links must take in order to traverse obstacles blocking their path by predicting the required angle. The control links are then actuated via PID control to climb the obstacle. The trained model is then used to generate data that is used as a benchmark to compare the efficiency of the proposed suspension type. \newline
\newline
Summary of the main contributions of the paper are as follows: 
\begin{itemize}
\item We developed and deployed a novel 5 bar suspension mechanism for our rover suspension that eliminates bogie overturn typically associated with the traditional rocker-bogie mechanism and its variants. Furthermore, the developed model outperforms the rocker-bogie model and provides greater stability during traversal.
\item We compared the active and passive variants of the proposed suspension model. We demonstrated that the active model outperforms the passive model on the chosen metrics that evaluate the suspension's effectiveness.
\item Instead of deploying traditional supervised methods, we demonstrated the use of deep Reinforcement Learning to train our agent. The latter addresses the inherent problems of the former, like abstracting huge amounts of labelled data before training and its inability to generalise to continuous action spaces.
\end{itemize}
\section{Related Works}
Brian H Wilcox et al. in \cite{wilcox2007athlete} detail the design of a Lunar Utility Vehicle called ATHLETE developed by NASA, Stanford University and Boeing Company. Several other works \cite{hidalgo2012kinematics}, \cite{cordes2014active} and \cite{cordes2018design} showcase different variants of the Sherpa rover suspension system. These designs (ATHLETE and Sherpa) utilise a highly actuated hybrid legged-wheeled suspension system that possesses higher mobility associated with active legged robots and wheeled robots' simplicity and energy efficiency. The suspension consists of drive wheels mounted at the extremities of an articulating leg with actuated joints. The system provides much greater modularity and redundancies than passive wheeled counterparts, including actively elevating wheels from the ground and actively altering the vehicle footprint and support polygon, which provides greater stability on slopes and enables the robot to fit into compact stow volumes. These proposed designs aim to address drawbacks typically associated with currently deployed wheeled rovers (e.g. NASA-JPL MER missions) and make a strong case for the use of actuated suspension systems.
Mohamed Krid and Faiz Benamar \cite{6094963} explore the development of a model predictive controller with a fast anti-roll suspension design. The suspension is suitable for rovers meant to travel at high speeds. However, the controller focuses only on minimising load transfer and energy consumption and has no provisions for easily climbing steep obstacles. \par
Recently, machine learning has been widely investigated in control problems and applied to vehicle suspension. For example, Ikbal Eski and Sahin Yildirim in \cite{eski2009vibration} compare two different control structures to control vehicle vibrations, displacements instigated by an active suspension model. They have developed a robust neural network and have compared its results to a fine-tuned PID controller. Simulations show the effectiveness of the proposed neural network model over the standard PID control system. The neural network control method designed in \cite{si2006neural}, \cite{eski2009vibration}makes full use of the fact that neural networks are suitable for the nonlinear system, control and achieve good results. However, as a supervised learning method, it requires the system to provide many samples with labels. Moreover, in generating the control strategy, only the current state is considered, but not the future state, which severely limits the method's utility.
The study by Mubin Khan \cite{finalkhan} focuses on devising a Reinforcement Learning algorithm applied to an active suspension control system. The algorithm aims to vary the model's damper coefficient to control the articulation of the axle shaft. Q learning is chosen to be the most appropriate algorithm for this study, thus making the model inefficient in extending to continuous action spaces. Furthermore, the environment is formulated as a fully observable Markov Decision Process that ignores the state uncertainty under the pretext of sensor readings being deemed 100 \% accurate and does not account for the stochastic nature of the agent's environment. \par 
 Ahmad Fares and Ahmad Bani Younes \cite{fares2020online} have used online Reinforcement Learning with the Temporal Difference (TD) advantage actor-critic to train an active suspension system controller. This model is compared to a controller proposed in \cite{konoiko2019deep} where a neural network was trained by the optimal PID and surpassed it under parameter uncertainties. The results showed that the trained Reinforcement Learning model obtains more optimal results under parameter uncertainty as compared to the supervised learning method proposed in \cite{konoiko2019deep}. Furthermore, the results encourage further studies by testing state-of-the-art continuous action Reinforcement Learning algorithms like the Deep Deterministic Policy Gradient (DDPG) \cite{lillicrap2019continuous} and Asynchronous Actor-Critic (A3C) \cite{mnih2016asynchronous}. 
Considering the continuity, stochasticity and complexity of the state and action space, Liu Ming et al. \cite{ming2020semi} base their semi-active suspension control system on the Deep Deterministic Policy Gradient (DDPG) algorithm \cite{lillicrap2019continuous}. The algorithm aims towards adjusting the active control force of a linear motor actuator, which further affects model parameters like body acceleration, displacement and the dynamic deflection of the suspension. The experimentation carried out in the time domain shows that the performance of the proposed model is much better than that of its passive counterpart when tested upon a randomly generated road profile. Mark N Howell et al. \cite{howell1997continuous} introduced a new Reinforcement Learning algorithm (CARLA) which focuses on generalising to continuous action spaces. The control objective of the algorithm is to minimise the mean squared acceleration of the vehicle body. The results indicate the proposed model to fair better than conventional passive models.\par
Passive suspension models have an inherent disadvantage of a delayed response since motion is induced due to forces that act on the suspension during traversal. These responses are unpredictable in nature and magnitude due to the stochasticity of the environment, which introduces increased vibrations and forces to the chassis and affects the stability by altering its configuration, namely the pitch. Integrating an active suspension model provides the robot with the ability to predict the response well in advance whilst interacting with the environment. The agent aims to perceive the environment variables and generate action space values, thus enabling the robot to emulate the response well in advance. Most supervised learning methods fail to capture the inherent stochasticity in the environment due to a lack of labelled data. Our work adopts a model-free Reinforcement Learning algorithm, Soft Actor-Critic (SAC), which models the environment accurately and generalises to perturbations in the environment without the necessity of collecting labelled data points. It also outperforms and converges faster than other model-free algorithms due to the use of a maximum entropy objective to generalise training to noises and perturbations in the environment thoroughly. 

\section{Simulation and Setup}
The environment for training the active suspension of the rover is simulated in Gazebo \cite{gazebo}. In addition, this environment has been incorporated with ROS  to ensure seamless communication and integration with all the robot's onboard sensors and controllers.  

\subsection{Unified Robot Description Format (URDF) Creation}
The rover's Solidworks robot model is exported into a universal robot description format file, or URDF, using the SW2URDF (Solidworks assembly to URDF exporter) plugin, which is available to the Solidworks users. A tree structure comprising parent link and child link connections is defined to simulate the five bar closed-chain mechanism appropriately. In addition, separate definitions are provided for joint types, joint positions, rotation axes, and joint limitations. The exporter's link properties configuration window is then used to configure the links' inertial properties to replicate real-world scenarios.

\subsection{Simulation Description Format (SDF) Creation}
The main downside of using a URDF file format is that it prevents the user from creating a closed-chain mechanism. So, the links forming the closed chain are detached in the URDF and reassembled in the SDF file.
We create the SDF by launching the robot's URDF in Gazebo, then editing and saving using the inbuilt model editor.
The generated model is then saved and stitched by hand, which closed the chain.

\begin{figure*}[htbp]
\begin{tabular}{cccc}
\subfloat[Before completing the closed chain (URDF)]{\includegraphics[width = 3.2in]{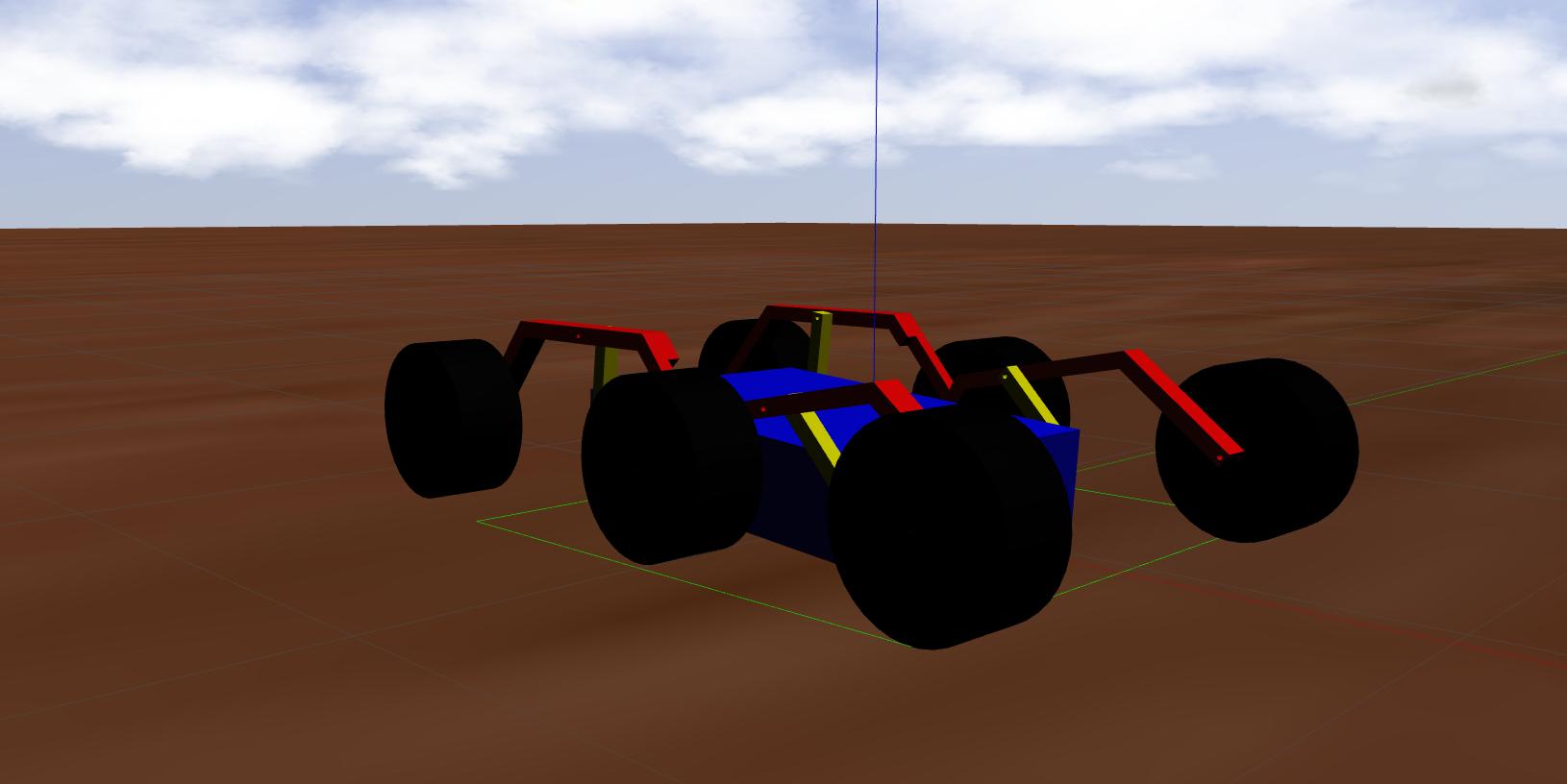}} &
\subfloat[After completing the closed chain (SDF)]{\includegraphics[width = 3.2in]{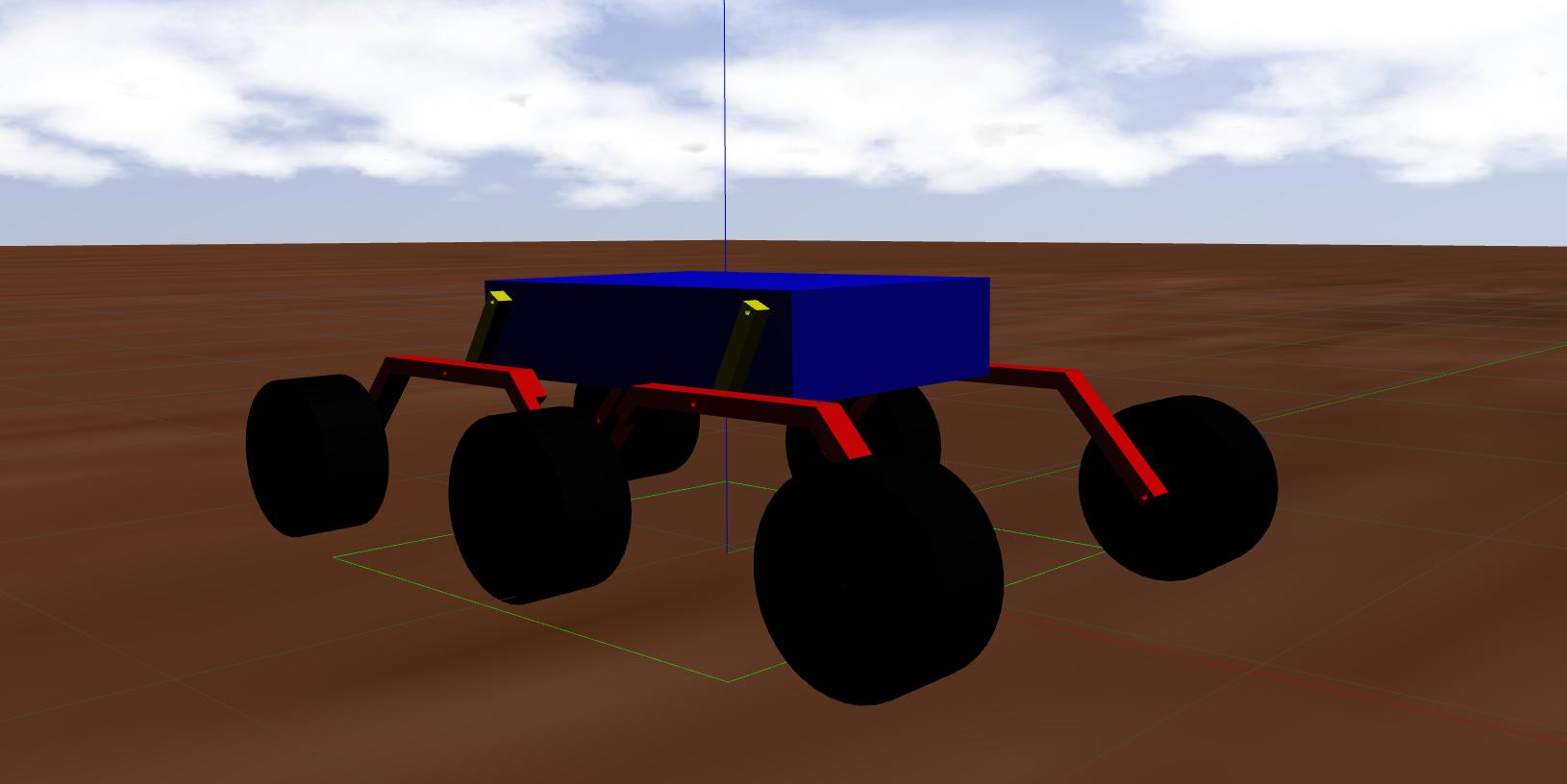}}\\
\end{tabular}
\caption{The rover simulated in Gazebo}
\label{gazebo rover}
\end{figure*} 

\subsection{Controllers and Plugins Added}
The multi-wheel hector differential drive plugin (\texttt{\url{http://wiki.ros.org/hector_gazebo_plugins}}) is used to teleoperate the robot. In addition, we have incorporated the Inertial Measurement Unit (IMU) plugin to record the Euler angle variations emerging from the robot.
A PID controller that has been manually tuned is attached to all the four motors of the control links.


\subsection{Training Environment}
gym-gazebo \cite{zamora2016extending} a wrapper for the OpenAI Gym \cite{gym} for Gazebo environments is used to build our training environment in the gym format. 
Our training environment consists of a randomly generated obstacle with a minimum height of 25 cm to a maximum height of 32 cm present in the robot's traversal path. The obstacle faces the rover with an edge perpendicular to the ground, as this scenario is the worst-case scenario for the rover to traverse. 
The obstacle is sufficient to be considered unavoidable, meaning the rover is forced to traverse it to reach its goal.

\subsection{Problem Description}\label{description}
The environment is formulated as a Markov Decision Process (MDP) to apply deep reinforcement learning. The notation used is presented as is from the Soft-Actor Critic paper \cite{pmlr-v80-haarnoja18b}. The MDP can be represented in tuple form as  \( \mathcal{(S, A, P, R)} \), where the state space \( \mathcal{S} \) and action space \( \mathcal{A} \) are continuous. The unknown state transition probability \( \mathcal{P}: \mathcal{S} \times \mathcal{S} \times \mathcal{A} \) $\rightarrow$ \( [0, \infty) \) represents the probability density of next state \( s_{t+1} \in \mathcal{S} \) given the current state \( s_t \in \mathcal{S} \) and action \( a_t \in \mathcal{A} \). The policy will be represented as \( \pi_\phi(a_t|s_t) \) where the policy represents the probability of taking action \( a_t \) at state \( s_t \). The environment emits a bounded reward \( \mathcal{R: S \times A} \) $\rightarrow$ \( [r_{min},r_{max}]\) at every transition.

The environment's observation space \( \mathcal{O} \) is a vector \( \mathcal{O} \in \mathbb{R}^{4} \). The observations available to the rover are as an array of form \( [pitch, roll, distance, height] \) with an upper bound of 50 and a lower bound of -50, which is obtained empirically. \( Pitch \) is defined as the angle subtended between the longitudinal axis of the chassis with the ground, while \( roll \) is defined as the angle subtended between the lateral axis of the chassis with the ground. The rover can learn a decent representation of its environment given these observations, which are part of the state space \( \mathcal{S} \). The \( pitch \) and \( roll \) of the rover at state \( s_t \) are obtained from an Inertial Measurement Unit. The \( distance \) is the distance of the rover from the face of the obstacle. \( height \) is the obstacle height as sourced from the environment. The action space \( \mathcal{A} \)  of the rover is another 4 dimensional vector \( \mathcal{A} \in \mathbb{R}^{4} \) in an array of form \( \mathit{[a_0, a_1, a_2, a_3]} \) with an upper bound of 1 and lower bound of -1. The angles \( \mathit{a_0, a_1} \) actuate the control links for the movement of the middle and rear pair. The angles \( \mathit{a_2,a_3} \) are used to actuate the control links of the two wheels approaching the obstacle. The angles are converted to the actual values in degrees by multiplying them by 37 (the maximum angle allowed to actuate the control link) before passing them to the ROS publishers for the joint position controllers.

Within the environment, when the rover reaches a certain threshold distance from the obstacle, the agent is made to start taking its steps. At every instance, the angles produced by the algorithm as part of its action space are enforced by the use of a PID controller provided in ROS control. The PID controller is tuned by hand to ensure no error in the angle achieved by the motor and the angle predicted by the agent.

\begin{figure*}[htbp]
\begin{tabular}{cccc}
\subfloat{\includegraphics[width = 2.1in]{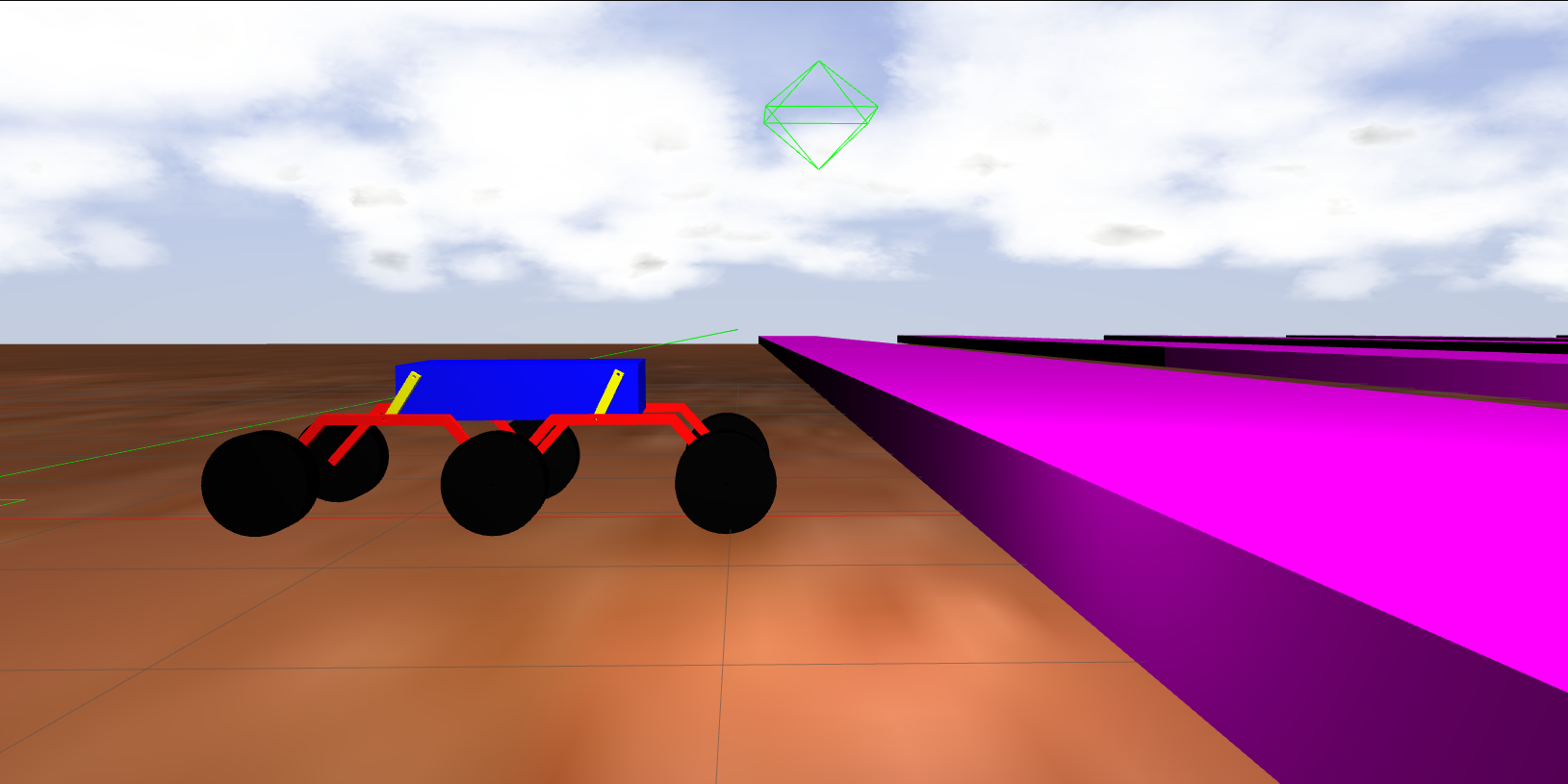}} &
\subfloat{\includegraphics[width = 2.1in]{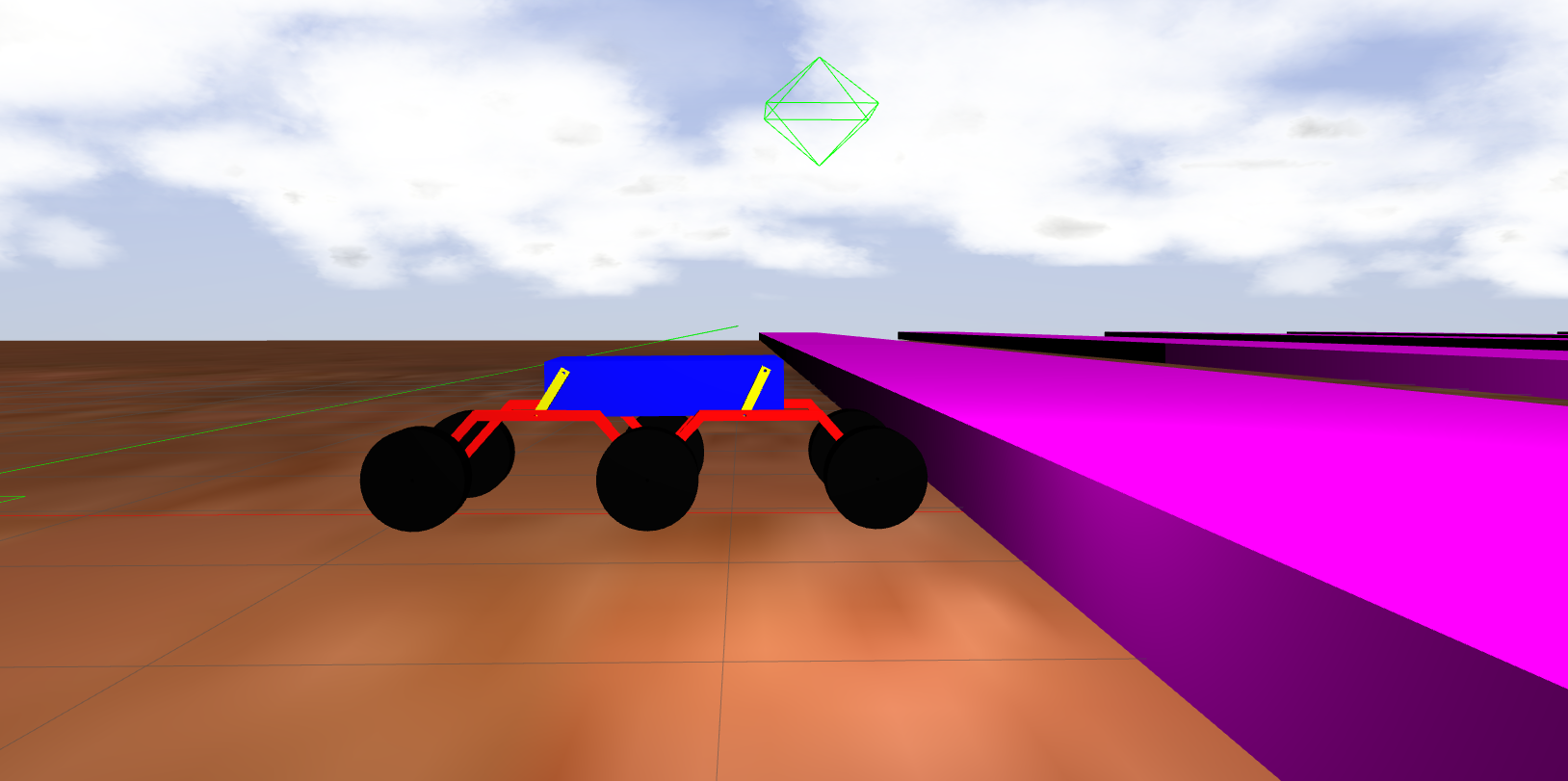}} &
\subfloat{\includegraphics[width = 2.1in]{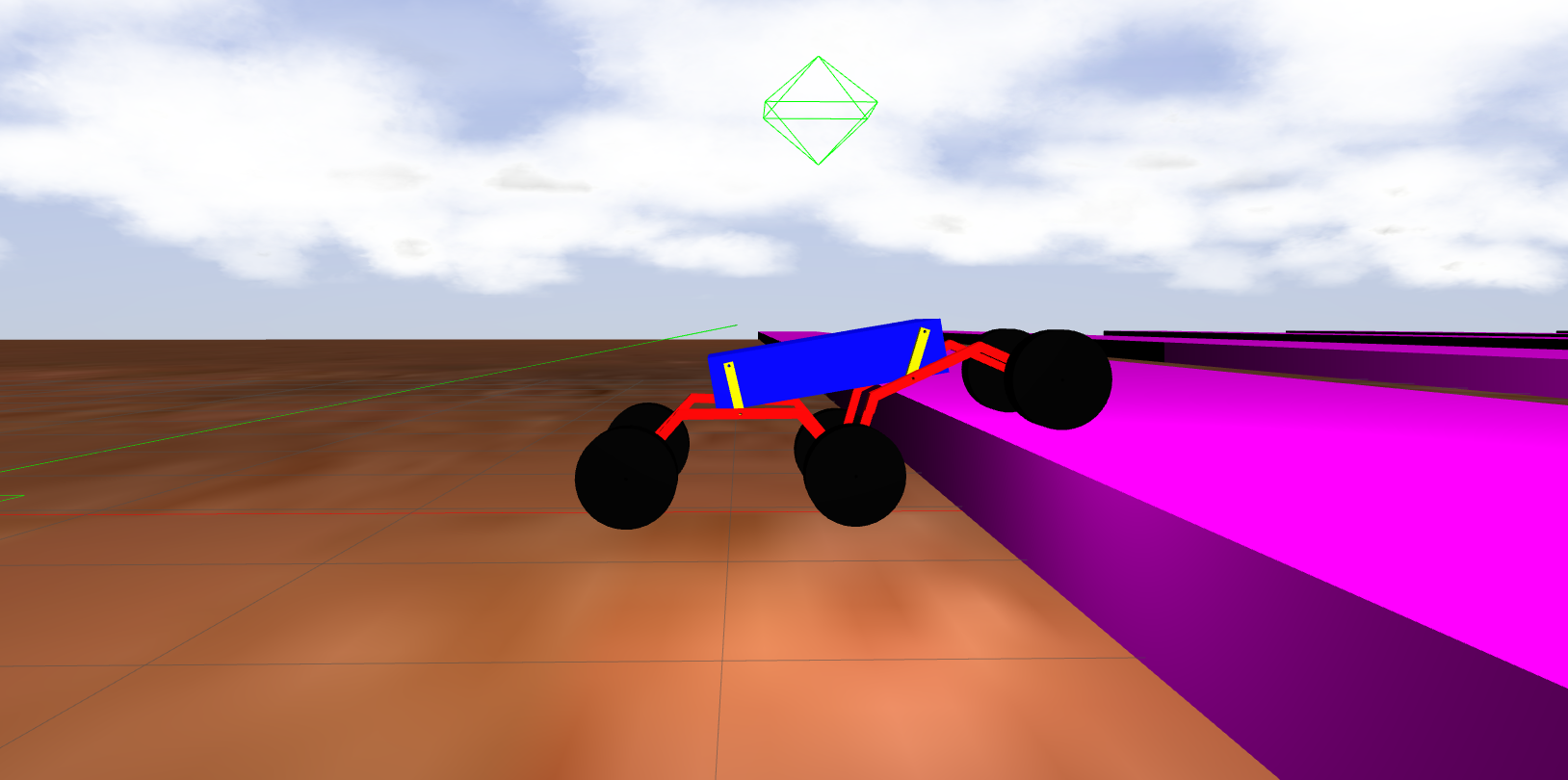}}\\
\subfloat{\includegraphics[width = 2.1in]{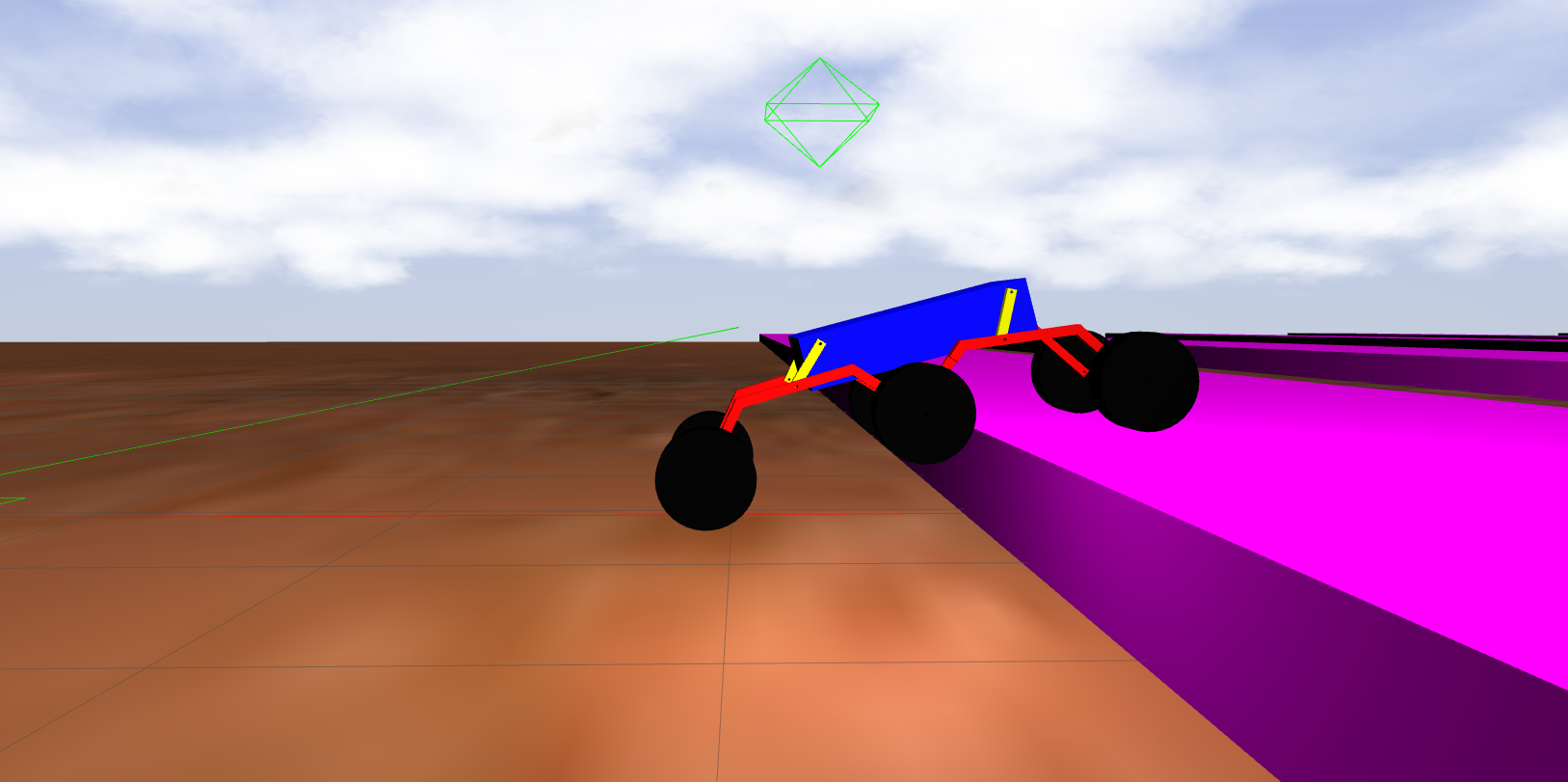}} &
\subfloat{\includegraphics[width = 2.1in]{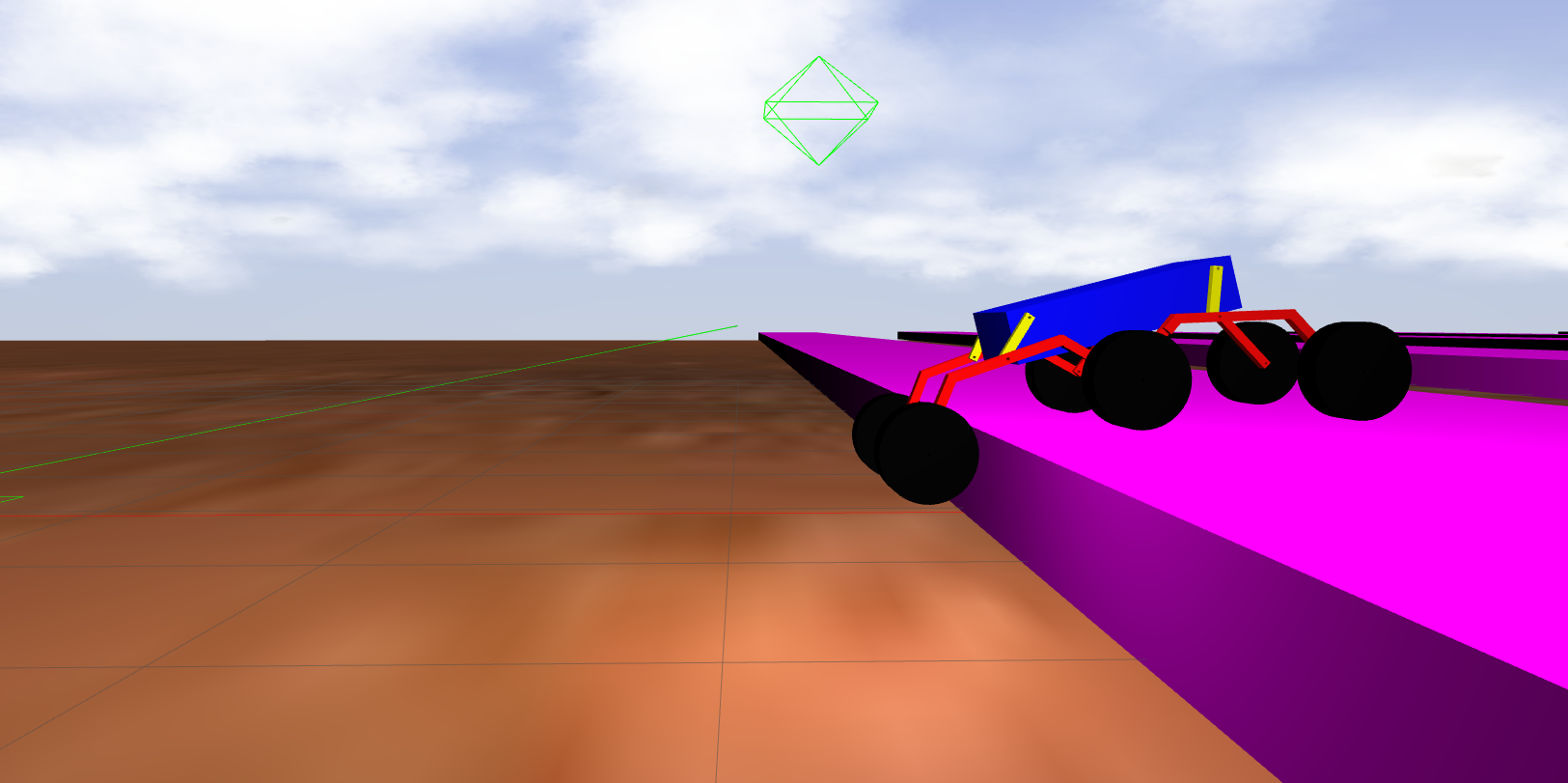}} &
\subfloat{\includegraphics[width = 2.1in]{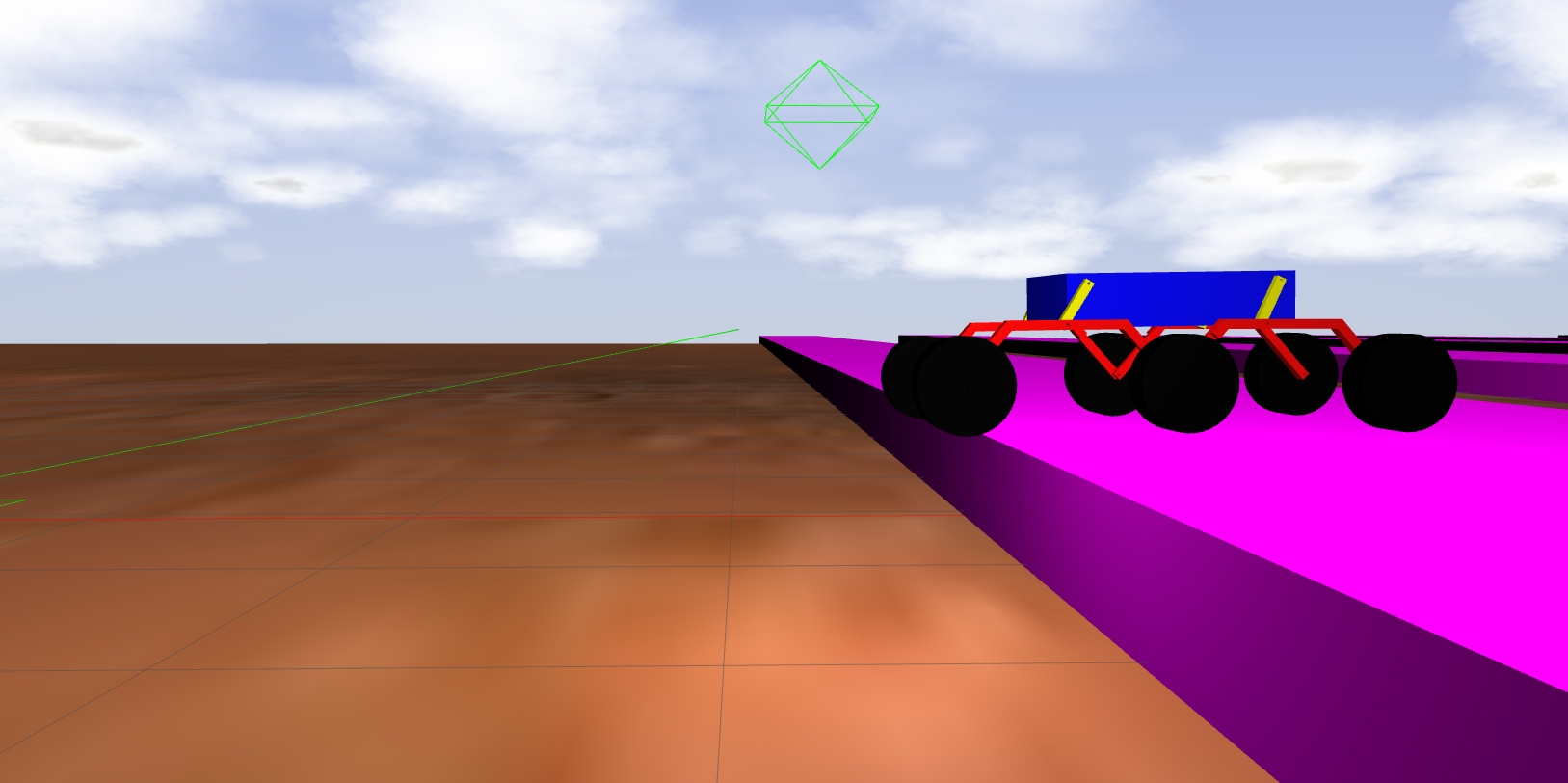}}\\
\end{tabular}
\caption{Trained model - SAC in action. The figure demonstrates how the trained model enables the rover to climb over a randomly generated obstacle of considerable size.}
\end{figure*}

\subsection{Soft Actor Critic (SAC)}
We have used the Soft-Actor Critic (SAC) \cite{haarnoja2019soft} algorithm for reinforcement learning. The choice of this algorithm is made based on the findings of Tuomas Haarnoja et al.\cite{haarnoja2019soft} showing that SAC generalises well to real-world robotics applications. 
The SAC algorithm aims to maximise a trade-off between two goals, the maximisation of reward and entropy, which is a measurement of the randomness of the policy. The use of the maximum entropy objective ensures the policy explores more widely while dropping unpromising avenues. Tuomas Haarnoja et al.\cite{haarnoja2019soft} and John Schulman et al. \cite{schulman2018equivalence} show that the use of this objective makes the model converge faster compared to the conventional method.
Thus the optimal policy aims to maximize the reward as well as the entropy at each visited state. This is given in Equation \ref{eq: Maxent_objective}
\begin{equation}
    \label{eq: Maxent_objective}
    \policy\opt = \arg\max_{\policy} \sum_{t} \E{(\st, \at) \sim \rho_\policy}{\reward(\st,\at) + \alpha\ent(\policy(\voidarg|\st))}
\end{equation}
where \( \alpha \) is the temperature term for entropy maximization that determines the importance of entropy maximization relative to the reward.
While for a complete derivation of the algorithm, we would encourage the readers to refer to the paper, in order to make the work self-contained, we will provide a look at the basic equations involved.

\begin{algorithm}[ht]
\caption{Soft Actor-Critic}
\label{alg:soft_actor_critic}
\hspace*{\algorithmicindent} \textbf{Input:} $\params_1$, $\params_2$, $\pparams$ \Comment{Initial parameters} \\
\hspace*{\algorithmicindent} \textbf{Output:} $\params_1$, $\params_2$, $\pparams$\Comment{Optimized parameters}
\begin{algorithmic}
\State $\bar \params_1 \leftarrow \params_1$, $\bar \params_2 \leftarrow \params_2$ \Comment{Initialize target network weights}
\State $\mathcal{D}\leftarrow\emptyset$ \Comment{Initialize an empty replay pool}
\For{each iteration}
	\For{each environment step}
	    \State $\at \sim \policy_\pparams(\at|\st)$ \Comment{Sample action from the policy}
	    \State $\stp \sim \pdyn(\stp| \st, \at)$ \Comment{Sample transition from the environment}
	    \State $\mathcal{D} \leftarrow \mathcal{D} \cup \left\{(\st, \at, \reward(\st, \at), \stp)\right\}$ \Comment{Store the transition in the replay pool}
	\EndFor
	\For{each gradient step}
	    \State $\params_i \leftarrow \params_i - \lambda_Q \hat \nabla_{\params_i} J_\Q(\params_i)$ for $i\in\{1, 2\}$ \Comment{Update the Q-function parameters}
	    \State $\pparams \leftarrow \pparams - \lambda_\policy \hat \nabla_\pparams J_\policy(\pparams)$\Comment{Update policy weights}
	    \State $\alpha \leftarrow \alpha - \lambda \hat \nabla_\alpha J(\alpha)$ \Comment{Adjust temperature}
	    \State $\bar\params_i\leftarrow \tau \params_i + (1-\tau)\bar\params_i$ for $i\in\{1,2\}$\Comment{Update target network weights}
	\EndFor
\EndFor
\end{algorithmic}
\end{algorithm}

SAC makes use of two soft Q functions. The soft Q function is calculated iteratively by applying the modified Bellman operator \( \mathcal{T}^\pi \) as 
\begin{equation}
    \mathcal{T}^\pi\mathcal{Q}(s_t, a_t)\triangleq\mathcal{R}(s_t, a_t)+\gamma\mathbb{E}_{s_{t+1}\backsim p}[V(s_{t+1})]
    \label{eq: Qfuncion}
\end{equation} 
where \( V(s_t) \) is the soft value function defined by
\begin{equation}
    V(\st) = \E{\at\sim\policy}{\Q(\st, \at) - \alpha\log\policy(\at|\st)}
\end{equation}
The policy has to be updated towards the exponential of the new soft Q function. A tractable policy is chosen which can correspond to a parameterized family of probability distributions. The information projection is defined in terms of the Kullback-Leibler Divergence. The policy obtained by the policy improvement step is updated by
\begin{equation}
    \pi_\mathrm{new} = \arg\underset{\pi'\in \Pi}{\min}\kl{\pi'(\centerdot|s_t)}{\frac{\exp\left(\frac{1}{\alpha}Q^{\pi_\mathrm{old}}(s_t, \centerdot)\right)}{Z^{\pi_\mathrm{old}}(s_t)}}
    \label{eq: KLdivergence}
\end{equation} where \( \alpha \) is the temperature parameter for entropy maximization and \( Z \) is a partition function which normalizes the distribuion but is ignored as it doesn't contribute to gradient calculations.
Several Objective functions are trained to optimise the parameters. The soft Q functions are parameters are trained by minimising the soft Bellman residual 
\begin{equation}
    J_\Q(\params) = \E{(\st, \at)\sim\mathcal{D}}{\frac{1}{2}\left(\Q_\params(\st, \at) - \left(\reward(\st, \at) + \discount \E{\stp\sim\pdyn}{V_{\bar\params}(\stp)}\right)\right)^2}
\end{equation}
This is optimized by the use of the stochastic gradient given in \ref{eq: theta_gradient} 
\begin{equation}
    \label{eq: theta_gradient}
    \hat \nabla_\params J_Q(\params) =  \nabla_\params \Q_\params(\at, \st) \left(\Q_\params(\st, \at) - \left(r(\st,\at) + \gamma \left(Q_{\bar\params}(\stp, \atp) - \alpha \log\left(\pi_\pparams(\atp|\stp\right)\right)\right)\right)
\end{equation}
The update uses the target Q function parameters $\bar\params$ calculated as the exponential moving average of the soft Q function weights.

The policy parameter \( \phi \) can be learnt by minimizing equation \ref{eq: KLdivergence}. In order to lower the variance estimate from the Q function, which is represented by a neural network and is differentiable, the reparameterization trick can be applied for the policy. Equation \ref{eq:reparameterization} shows the reparameterization, 
\begin{equation}
    \at = f_\pparams(\epsilon_t; \st),
    \label{eq:reparameterization}
\end{equation}
where \( \epsilon_t \) is a noise vector sampled from a fixed distribution such as a Gaussian. The final objective function for the parameter \( \phi \) is given by
\begin{equation}
    J_\policy(\pparams) = \E{\st\sim\mathcal{D},\epsilon_t\sim\gauss}{\alpha \log \policy_\pparams(f_\pparams(\epsilon_t;\st)|\st) - Q_\params(\st, f_\pparams(\epsilon_t;\st))},
\label{eq:reparam_objective}
\end{equation} where \( \pi_\phi \) is reparameterized in terms of \( f_\phi \).
The stochastic gradient can be approximated with \ref{eq: policy_gradient}
\begin{equation}
    \label{eq: policy_gradient}
    \hat\nabla_\pparams J_\policy(\pparams) = \nabla_\pparams \alpha \log\left( \policy_\pparams(\at|\st)\right) + (\nabla_\at \alpha \log \left(\policy_\pparams(\at|\st)\right)
- \nabla_\at Q(\st, \at))\nabla_\pparams f_\pparams(\epsilon_t;\st)
\end{equation}

The final parameter to be tuned is the temperature for the maximization of entropy. To get the optimal temperature \( \alpha \) the dual equation in equation \ref{eq: dual} must be solved.
\begin{equation}
    \label{eq: dual}
    \alpha_t\opt = \arg \min_{\alpha_t} \E{\at\sim \policy_t\opt}{- \alpha_t\log\pi_t\opt(\at|\st; \alpha_t) - \alpha_t \bar\ent}
\end{equation}
Solving the recursive equation for this dual problem would give us the optimal temperature, but we use function approximators like neural networks and gradient descent in practice. Dual Gradient Descent is used by \cite{haarnoja2019soft} to solve the objective in \ref{eq:ecsac:alpha_objective} to get the optimal temperature.
\begin{equation}
    J(\alpha)  = \E{\at\sim \policy_t}{ - \alpha\log\pi_t(\at|\st) - \alpha \bar\ent}.
\label{eq:ecsac:alpha_objective}
\end{equation}
These objectives form the core of the SAC algorithm, and the parameters are approximated via stochastic gradient descent.
The minimum of the two soft Q functions is used to calculate the stochastic gradient in \ref{eq: theta_gradient} and the policy gradients in \ref{eq: policy_gradient} in order to minimise the positive biases in the policy improvement step \cite{haarnoja2019soft}. This is also known to speed up training on harder tasks significantly. The final algorithm for SAC is given in algorithm \ref{alg:soft_actor_critic}.

For our Gazebo simulation, we have compared SAC to similar off-policy algorithms such as Deep Delayed Deterministic Policy Gradients (DDPG)  \cite{lillicrap2019continuous} and Twin Delayed Deep Deterministic Policy Gradients (TD3) \cite{pmlr-v80-fujimoto18a}. It has also been compared to a popular on-policy learning algorithm, Proximal Policy Optimization (PPO)  \cite{schulman2017proximal}.

\subsection{Reward Function}
The rewards obtained by the agent are incredibly crucial to the learning process. Rewards indicate the quality of actions chosen by the agent in reaching the next state. Positive rewards encourage the agent to accumulate as much reward as possible, while negative rewards encourage the agent to reach the desired state quickly. The reward for the rover at every step \( s_t \) is defined by algorithm \ref{alg: rewards}.
\begin{algorithm}
\caption{The Reward function for every step}
\label{alg: rewards}
\hspace*{\algorithmicindent} \textbf{Input:} $s_{t+1}$ \\
\hspace*{\algorithmicindent} \textbf{Output:} $r$
\begin{algorithmic}
    \State Episode End = $False$;
    \State Reward \( r \)=0;
    \If{ $pitch$ > 20\degree}
        \State \( r \)=-100;
        \State Episode End = $True$;
    \EndIf
    \If{$yaw$> 10\degree}
        \State \( r \)=-100;
        \State Episode End = $True$;
    \EndIf
    \If{Obstacle is crossed}
        \State \( r \)=100;
        \State Episode End = $True$;
    \EndIf
    \If{Obstacle is not crossed and time steps>430}
        \State \( r \)=-50;
        \State Episode End = $True$;
    \EndIf
\end{algorithmic}
\end{algorithm}
This simple reward formulation penalises the rover for veering off course while attempting to climb or failing to cross the obstacle fast enough. If the rover successfully crosses the obstacle, it gets a single positive reward, and the next episode begins. In line with the rover's objective to maintain the chassis' stability, a negative reward is provided if the rover is tilted beyond a threshold and the episode ends. No intermediate rewards were required in the training process to guiding the policy, and a single positive reward on reaching the goal helps the model converge fastest.

\subsection{Software and Hardware}
The environment has been created from scratch by the authors. All the algorithms make use of the Stable Baselines framework for their implementation. PyTorch is used as the Reinforcement Learning framework, and all the programs are implemented in Python. The programs were developed on Ubuntu 20.04. Multiple training runs were conducted with each algorithm, all resulting in convergence. All training runs were conducted on a Lenovo ThinkPad L-470 with a quadcore i5-7200 CPU and 7.9 GB RAM. The codebase for the environment and agent can be found in {\texttt{\url{https://github.com/Mars-Rover-Manipal/Active-Suspension}}}. 

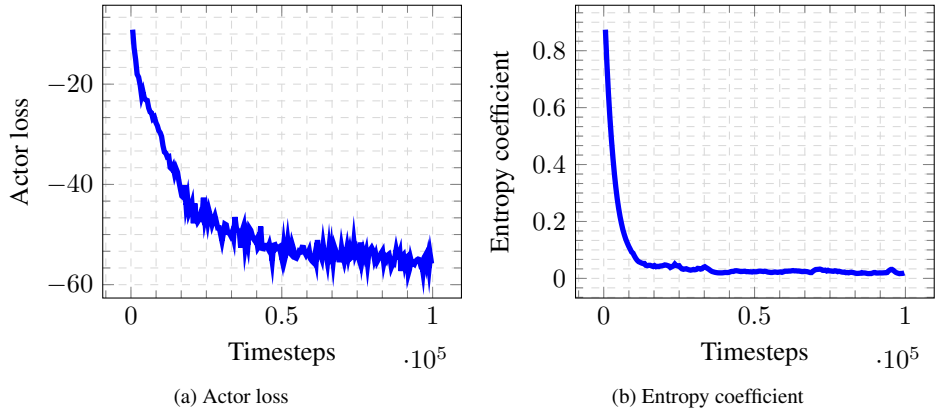
\begin{figure*}[!h]
\begin{center}
    \subfloat[Actor loss]{
    \begin{tikzpicture}
      \begin{axis}[
          width=2.5 in, 
          grid=both, 
          grid style={dashed,gray!30}, 
          minor tick num=5,
          xlabel=Timesteps , 
          ylabel= Actor loss]
        \addplot+ [line width=2pt,mark=none] 
        table[x= Step,y= Value, col sep=comma] {run-SAC_5-tag-train_actor_loss.csv};
      \end{axis}
    \end{tikzpicture}
}
\hspace{0cm}
    \subfloat[Entropy coefficient]{
    \begin{tikzpicture}
      \begin{axis}[
          width=2.5 in, 
          grid=both, 
          grid style={dashed,gray!30}, 
          minor tick num=5,
          xlabel=Timesteps , 
          ylabel= Entropy coefficient]
        \addplot+ [line width=2pt,mark=none] 
        table[x= Step,y= Value, col sep=comma] {run-SAC_5-tag-train_ent_coef.csv};
      \end{axis}
    \end{tikzpicture}
    }
\caption{Training Plots of SAC obtained after being trained on the \texttt{GazeboMarsLsd-v0} environment}
\label{fig: trainingresults}
\end{center}
\end{figure*}
\section{Results and Discussion}
Figure \ref{fig: trainingresults} shows the convergence of the model as the actor loss and the entropy coefficient \( \alpha \) both decrease as the model converges to its optimal policy. Furthermore, \( \alpha \) starts to approach zero as the model finds the optimal policy indicating a lesser need for exploration and more focus on maximising the reward.
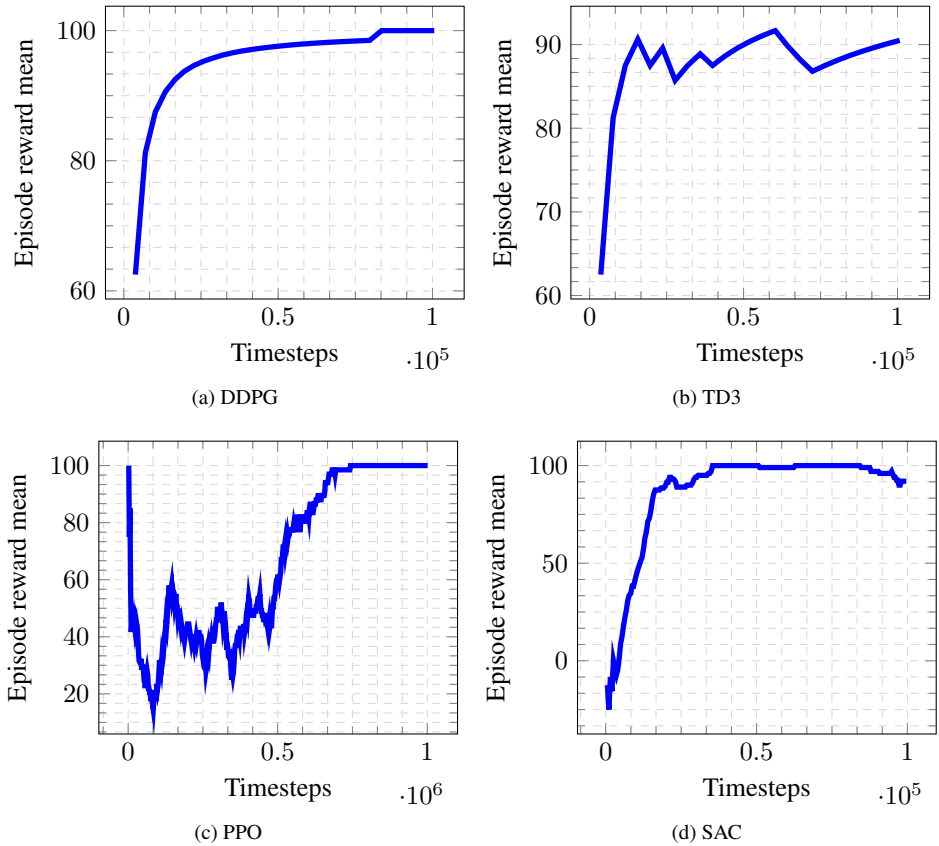
\begin{figure}[!htbp]
\begin{center}
    \subfloat[DDPG]{
    \begin{tikzpicture}
      \begin{axis}[
          width=2.5 in, 
          grid=both, 
          grid style={dashed,gray!30}, 
          minor tick num=5,
          xlabel=Timesteps , 
          ylabel= Episode reward mean]
        \addplot+ [line width=2pt,mark=none] 
        table[x= Step,y= Value, col sep=comma] {run-DDPG_1-tag-rollout_ep_rew_mean.csv};
      \end{axis}
    \end{tikzpicture}
}
\hspace{0cm}
    \subfloat[TD3]{
    \begin{tikzpicture}
      \begin{axis}[
          width=2.5 in, 
          grid=both, 
          grid style={dashed,gray!30}, 
          minor tick num=5,
          xlabel=Timesteps , 
          ylabel= Episode reward mean]
        \addplot+ [line width=2pt,mark=none] 
        table[x= Step,y= Value, col sep=comma] {run-TD3_3-tag-rollout_ep_rew_mean.csv};
      \end{axis}
    \end{tikzpicture}
    }
\hspace{0cm}
    \subfloat[PPO]{
    \begin{tikzpicture}
      \begin{axis}[
          width=2.5 in, 
          grid=both, 
          grid style={dashed,gray!30}, 
          minor tick num=5,
          xlabel=Timesteps , 
          ylabel= Episode reward mean]
        \addplot+ [line width=2pt,mark=none] 
        table[x= Step,y= Value, col sep=comma] {run-PPO_2-tag-rollout_ep_rew_mean.csv};
      \end{axis}
    \end{tikzpicture}
    }
\hspace{0cm}
    \subfloat[SAC]{
    \begin{tikzpicture}
      \begin{axis}[
          width=2.5 in, 
          grid=both, 
          grid style={dashed,gray!30}, 
          minor tick num=5,
          xlabel=Timesteps , 
          ylabel= Episode reward mean]
        \addplot+ [line width=2pt,mark=none] 
        table[x= Step,y= Value, col sep=comma] {run-SAC_5-tag-rollout_ep_rew_mean.csv};
      \end{axis}
    \end{tikzpicture}
}
\caption{Mean episode rewards plotted as a function of timesteps}
\label{fig:eprew}
\end{center}
\end{figure}

The total average return of evaluation rollouts
during training, i.e. the mean reward obtained for progressive episodes and the mean episode length is plotted progressively for 1M timesteps for PPO and 100k timesteps for DDPG, SAC and TD3. SAC learns considerably faster than PPO due to the large batch sizes PPO needs to learn stably on more high-dimensional and complex tasks \cite{haarnoja2019soft}. The plots in Figure \ref{fig:eprew} depicts that SAC converges to an optimal stochastic policy in much fewer timesteps as compared to other tested on and off policy baselines. 
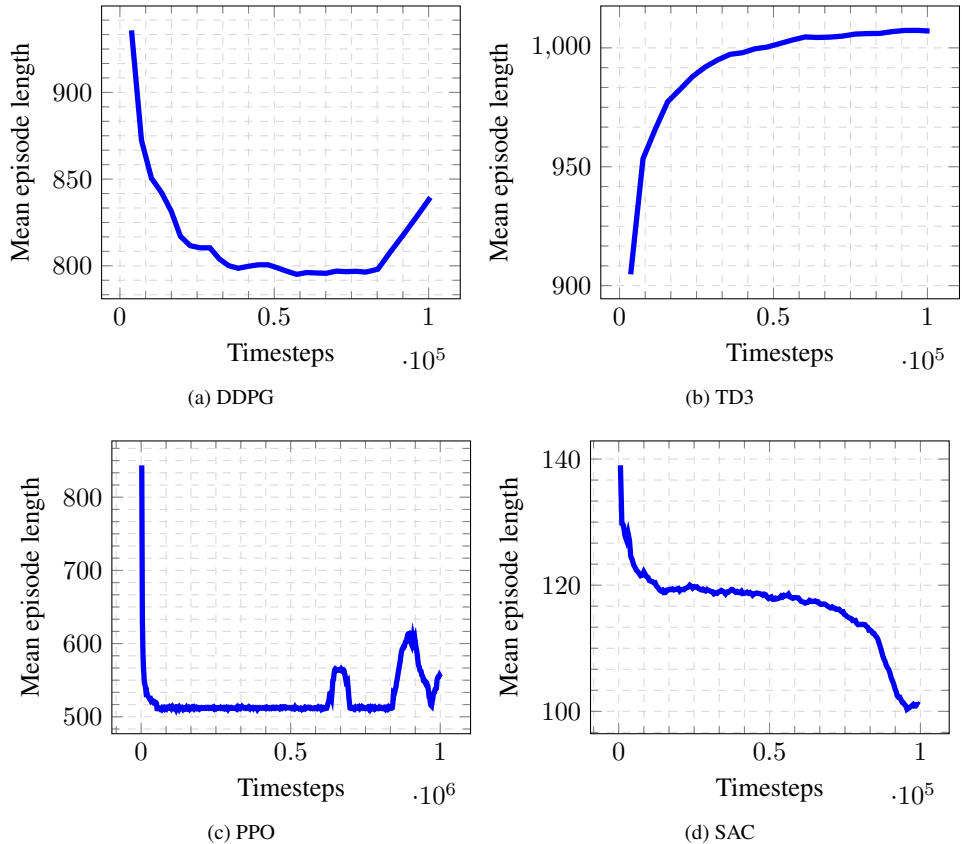
\begin{figure}[!htbp]
\begin{center}
    \subfloat[DDPG]{
    \begin{tikzpicture}
      \begin{axis}[
          width=2.5 in, 
          grid=both, 
          grid style={dashed,gray!30}, 
          minor tick num=5,
          xlabel=Timesteps , 
          ylabel= Mean episode length]
        \addplot+ [line width=2pt,mark=none] 
        table[x= Step,y= Value, col sep=comma] {run-DDPG_1-tag-rollout_ep_len_mean.csv};
      \end{axis}
    \end{tikzpicture}
}
\hspace{0cm}
    \subfloat[TD3]{
    \begin{tikzpicture}
      \begin{axis}[
          width=2.5 in, 
          grid=both, 
          grid style={dashed,gray!30}, 
          minor tick num=5,
          xlabel=Timesteps , 
          ylabel= Mean episode length]
        \addplot+ [line width=2pt,mark=none] 
        table[x= Step,y= Value, col sep=comma] {run-TD3_3-tag-rollout_ep_len_mean.csv};
      \end{axis}
    \end{tikzpicture}
    }
\hspace{0cm}
    \subfloat[PPO]{
    \begin{tikzpicture}
      \begin{axis}[
          width=2.5 in, 
          grid=both, 
          grid style={dashed,gray!30}, 
          minor tick num=5,
          xlabel=Timesteps , 
          ylabel= Mean episode length]
        \addplot+ [line width=2pt,mark=none] 
        table[x= Step,y= Value, col sep=comma] {run-PPO_2-tag-rollout_ep_len_mean.csv};
      \end{axis}
    \end{tikzpicture}
    }
\hspace{0cm}
    \subfloat[SAC]{
    \begin{tikzpicture}
      \begin{axis}[
          width=2.5 in, 
          grid=both, 
          grid style={dashed,gray!30}, 
          minor tick num=5 ,
          xlabel=Timesteps , 
          ylabel= Mean episode length]
        \addplot+ [line width=2pt,mark=none] 
        table[x= Step,y= Value, col sep=comma] {run-SAC_5-tag-rollout_ep_len_mean.csv};
      \end{axis}
    \end{tikzpicture}
}
\caption{Mean episode lengths plotted as a function of timesteps. SAC shows the least mean episode length amongst the other aforementioned algorithms.}
\label{fig:eplen}
\end{center}
\end{figure}
Figure \ref{fig:eplen} depicts that SAC also exhibits the least mean episode length amongst all trained algorithms and its successive convergence through learning as opposed to TD3 and DDPG. Overall, this implies that SAC outperforms the other baselines models for our provided task with a large margin. The results above suggest that algorithms based on the maximum entropy principle can outperform conventional Reinforcement Learning methods on challenging tasks \cite{haarnoja2019soft}.  \par
\begin{figure}[!htbp]
  \begin{center}
    \begin{tikzpicture}
      \begin{axis}[
          width=2.5 in, 
          grid=both, 
          grid style={dashed,gray!30}, 
          minor tick num=5,
          xlabel=Time , 
          ylabel= Pitch,
          x unit=\si{\sec}, 
          y unit=\si{\degree},
          legend style={at={(1.3, 1)},anchor=north}
        ]
        \addplot+ [line width=2pt, mark=none] 
        table[x= Time (sec),y= Passive, col sep=comma] {Pitch.csv}; 
        \addlegendentry{Passive}
        \addplot+ [line width=2pt, mark=none]  
        table[x= Time (sec),y= Active, col sep=comma] {Pitch.csv}; 
        \addlegendentry{Active}
      \end{axis}
    \end{tikzpicture}
    \caption{The plot depicts the chassis pitch variation with respect to time for active and passive suspension model whilst traversing over an obstacle.}
    \label{fig:pitch_active_v_passive}
  \end{center}
\end{figure}
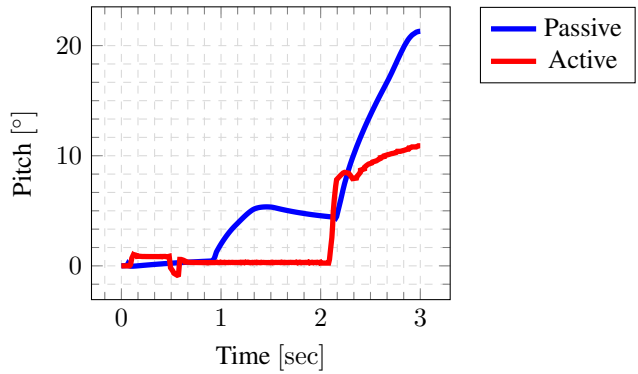

As aforementioned, chassis pitch is the angle subtended between the longitudinal axis of the chassis and the plane of the ground. The pitch of the chassis must is minimised in order to maintain vehicle stability. Toppling occurs when the projection of the centre of gravity of the vehicle on the ground moves outside the vehicle's support polygon \cite{barlas2004design}. High chassis pitch increases the risks of toppling, especially on inclined surfaces. Frequent fluctuation in pitch also causes scientific components and instruments to experience inertial forces, which is highly undesirable.
Transient responses of both passive and active suspension system are determined in the time domain. Time domain analysis is carried out for a step profile of a height of 32 cm. The suspension and the terrain primarily determine the nature of variation of the pitch. From Fig \ref{fig:pitch_active_v_passive}, the peak amplitude of the pitch is 21.31\degree  for the passive system, whereas, for the active system, it is 10.73\degree. The peak amplitude is reduced by a magnitude of 10.58\degree, demonstrating the effectiveness of the proposed active suspension in maintaining chassis stability, which is the singular most important factor in suspension design. \par

\begin{figure}[!htbp]

 \begin{center}
    \begin{tikzpicture}
      \begin{axis}[
          width=2.5 in, 
          grid=both, 
          grid style={dashed,gray!30}, 
          minor tick num=5,
          xlabel=Timesteps , 
          ylabel= Velocity,
          x unit=, 
          y unit=m/sec,
          legend style={at={(1.3, 1)},anchor=north}
        ]
        \addplot+ [line width=2pt,mark=none] 
        table[x= Timesteps,y= Passive, col sep=comma] {Linear-Velocity.csv}; 
        \addlegendentry{Passive}
        \addplot+ [line width=2pt, mark=none]  
        table[x= Timesteps,y= Active, col sep=comma] {Linear-Velocity.csv}; 
        \addlegendentry{Active}
      \end{axis}
    \end{tikzpicture}
    \caption{Travel velocity of the model plotted as a function of time.}
    \label{fig:velocity_active_v_passive}
  \end{center}

\end{figure}
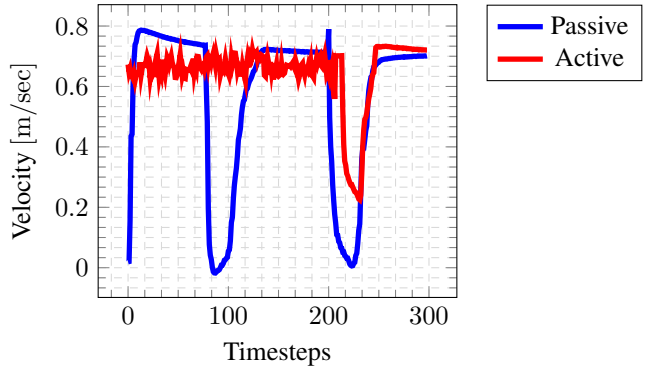

The travel velocity is the velocity of the rover during its course of travel over an obstacle. Figure \ref{fig:velocity_active_v_passive} is obtained by plotting the travel velocity for both the active and passive models for a 32 cm step obstacle. In a passive suspension system, the motion of the suspension mechanism is induced purely in response to terrain variations. Thus, while traversing the obstacle as each wheel engages, the mechanism idles as reaction forces need to reach a threshold value to cause the suspension mechanism to articulate. Hence the passive system demonstrates lower velocity values while traversing the obstacle than the active counterpart proposed in this paper, which exhibits no time lag as it can preemptively actuate the mechanism in response to perceived terrain undulations.
Furthermore, the velocity plot (see Fig \ref{fig:velocity_active_v_passive}) corresponding to the passive model indicates successive rise and fall in the velocity as each wheel engages with the vertical face of the step. On the other hand, the active model succeeds in maintaining an almost constant velocity profile while encountering the step due to its ability to predict the motor angles well before encountering it. The plot shows that the proposed active model fares better than its passive counterpart in maintaining its constant travel velocity of 0.7 m/s (chosen for the simulation empirically). The negative portion of the graph in the velocity profile of the passive model corresponds to the rebound the rover experiences when the wheels hit the obstacle. This rebound being absent in the active model makes traversal smoother and faster.   
\section{Conclusion}
The work proposes a novel five bar suspension architecture that eliminates the bogie overturn problem, an inherent disadvantage present in the conventional rocker-bogie suspension model. Furthermore, the adaptation of an active model has proven advantageous compared to its passive counterpart as it minimises the chassis pitch fluctuations greatly and succeeds in maintaining a nearly constant velocity during its course of travel. We have presented an end to end learning system for the automation of active suspensions for Mars rovers that use a partially observed environment to control the actuation of the suspension control links. Our experiments converged on a policy for traversing unavoidable obstacles within 4 hours of training time without using a powerful GPU. The algorithm, Soft Actor-Critic (SAC), based on an entropy constrained Reinforcement Learning objective outperforms other model-free Reinforcement Learning methods in our environment. However, we obtained the current results through simulation studies. Whether the practical application has similar effects still needs our further validation. Future works also include integrating the current learning architecture with an efficient path planning algorithm and multi-task learning of obstacle avoidance and traversal.

\section{Acknowledgements}
We would like to thank Mars Rover Manipal for providing us with the necessary resources to complete the research.

\bibliographystyle{plain}
\bibliography{custom}
\end{document}

%% file: defs.tex
\newcommand{\E}[2]{\operatorname{\mathbb{E}}_{#1}\left[#2\right]}

\newcommand{\density}{p}

\newcommand{\kl}[2]{\mathrm{D_{KL}}\left(#1\;\middle\|\;#2\right)}

\newcommand{\ent}{\mathcal{H}}

\newcommand{\voidarg}{{\,\cdot\,}}

\newcommand{\state}{\mathbf{s}}

\newcommand{\st}{{\state_t}}

\newcommand{\stp}{{\state_{t+1}}}

\newcommand{\pdyn}{\density}

\newcommand{\action}{\mathbf{a}}

\newcommand{\at}{{\action_t}}

\newcommand{\atp}{{\action_{t+1}}}

\newcommand{\opt}{^*}

\newcommand{\reward}{\mathcal{R}}

\newcommand{\Q}{Q}

\newcommand{\policy}{\pi}

\newcommand{\params}{\theta}

\newcommand{\pparams}{{\phi}}   

\newcommand{\gauss}{\mathcal{N}}

\newcommand{\discount}{\gamma}